\documentclass[9pt, conference]{IEEEtran}
\IEEEoverridecommandlockouts
\usepackage{cite}
\usepackage{amsmath,amssymb,amsfonts}
\usepackage{graphicx}
\usepackage{textcomp}
\usepackage{xcolor}
\usepackage{algorithm}
\usepackage{algpseudocode} 
\usepackage{booktabs}
\usepackage{multirow}
\usepackage{url}
\def\BibTeX{{\rm B\kern-.05em{\sc i\kern-.025em b}\kern-.08em
    T\kern-.1667em\lower.7ex\hbox{E}\kern-.125emX}}
\begin{document}
\fontsize{10pt}{12pt}\selectfont

\title{MathReader : Text-to-Speech for Mathematical Documents}

\author{
    \IEEEauthorblockN{
	Sieun Hyeon\textsuperscript{1}, 
	Kyudan Jung\textsuperscript{2}, 
        Nam-Joon Kim\textsuperscript{1}\IEEEauthorrefmark{2},
        Hyun Gon Ryu\textsuperscript{3},
        Jaeyoung Do\textsuperscript{1, 4}\IEEEauthorrefmark{2}
    }\thanks{\IEEEauthorrefmark{2}: Corresponding authors}
\IEEEauthorblockA{\textit{\textsuperscript{1}Department of Electrical and Computer Engineering, Seoul National University}}
\IEEEauthorblockA{\textit{\textsuperscript{2}Department of Mathematics, Chung-Ang University}}
\IEEEauthorblockA{\textit{\textsuperscript{3}NVIDIA}}
\IEEEauthorblockA{\textit{\textsuperscript{4}Interdisciplinary Program in Artificial Intelligence, Seoul National University}}
\IEEEauthorblockA{zxc2692@snu.ac.kr , wjdrbeks1021@cau.ac.kr , knj01@snu.ac.kr , hryu@nvidia.com , jaeyoung.do@snu.ac.kr
}
}

\maketitle

\begin{abstract}
TTS (Text-to-Speech) document reader from Microsoft, Adobe, Apple, and OpenAI have been serviced worldwide. They provide relatively good TTS results for general plain text, but sometimes skip contents or provide unsatisfactory results for mathematical expressions. This is because most modern academic papers are written in LaTeX, and when LaTeX formulas are compiled, they are rendered as distinctive text forms within the document. However, traditional TTS document readers output only the text as it is recognized, without considering the mathematical meaning of the formulas. To address this issue, we propose MathReader, which effectively integrates OCR, a fine-tuned T5 model, and TTS. MathReader demonstrated a lower Word Error Rate (WER) than existing TTS document readers, such as Microsoft Edge and Adobe Acrobat, when processing documents containing mathematical formulas. MathReader reduced the WER from 0.510 to 0.281 compared to Microsoft Edge, and from 0.617 to 0.281 compared to Adobe Acrobat. This will significantly contribute to alleviating the inconvenience faced by users who want to listen to documents, especially those who are visually impaired. The code is available
at https://github.com/hyeonsieun/MathReader.
\end{abstract}
\section{Introduction}

When reading documents, we often find ourselves in situations where we must rely on hearing rather than sight. This is especially true for visually impaired individuals as they cannot directly perceive text documents with their eyes and need someone to read them aloud. In such cases, a TTS document reader can be used. This technology provides the ability to automatically read text within documents, and for visually impaired individuals, it is essential to understand the content.

Many current software applications provide these services. For example, programs such as Microsoft Edge\cite{microsoftedge} and Adobe Acrobat\cite{adobe} offer a \textit{read aloud}  feature for documents saved in the PDF format. However, this functionality is not perfect, and we found that both programs output inaccurate speech when reading the formulas. This poses a significant barrier for visually impaired individuals trying to understand documents in which formulas, such as mathematics or physics, play a crucial role.

The main reason formulas are read inaccurately is that, in many modern academic documents, these formulas are typically written in LaTeX and then compiled. As a result, the formulas are displayed in a distinct format that is different from regular text. However, traditional TTS systems\cite{pmlr-v139-kim21f, NEURIPS2023_2d8911db, NEURIPS2023_58d0e78c, kong23_interspeech, 10409539, shen2023naturalspeech} read these formulas as simple texts without considering their mathematical meaning, leading to inaccurate speech that does not reflect the true meaning. Additionally, older documents were not written in LaTeX and only existed as images, which prevents the text from being recognized by TTS systems and thus results in the formulas not being read at all.

One solution to this problem is to convert all documents uniformly into markup text files. For this, we can use OCR (Optical Character Recognition)\cite{wang2020pdf2latex, blecher2023nougatneuralopticalunderstanding, peng2023upocr, orji2023advancing, mishra2018sequence}. However, the text extracted from the document through OCR is not general English but LaTeX code containing special symbols. When this is directly input into a TTS system, general English without special symbols is converted into speech correctly, but the formula parts containing special symbols such as \textbackslash, [, and = are not output as speech. In other words, the formula parts extracted from the document cannot be converted into speech using a general TTS model. Therefore, we added a process that converted the LaTeX parts representing formulas into spoken English before passing the OCR results to the TTS system. An example of converting an LaTeX formula into spoken English is presented in Table 1.

\begin{table}[h]
\centering
\small
\caption{
An example of converting a formula into Spoken English.}
\begin{tabular}{@{}p{1.45cm}|p{6.5cm}}
    \toprule
    \textbf{formula}  & We found that $ e^{ix} = \cos(x) + i\sin(x) $ is important.\\
    \midrule
    \textbf{Spoken English} & We found that 
 \textit{e to the power of i x equals cosine of x plus i sine of x} is important.\\
    \bottomrule
\end{tabular}
\end{table}

\begin{figure*}[t]
\centering
\includegraphics[width=\textwidth]{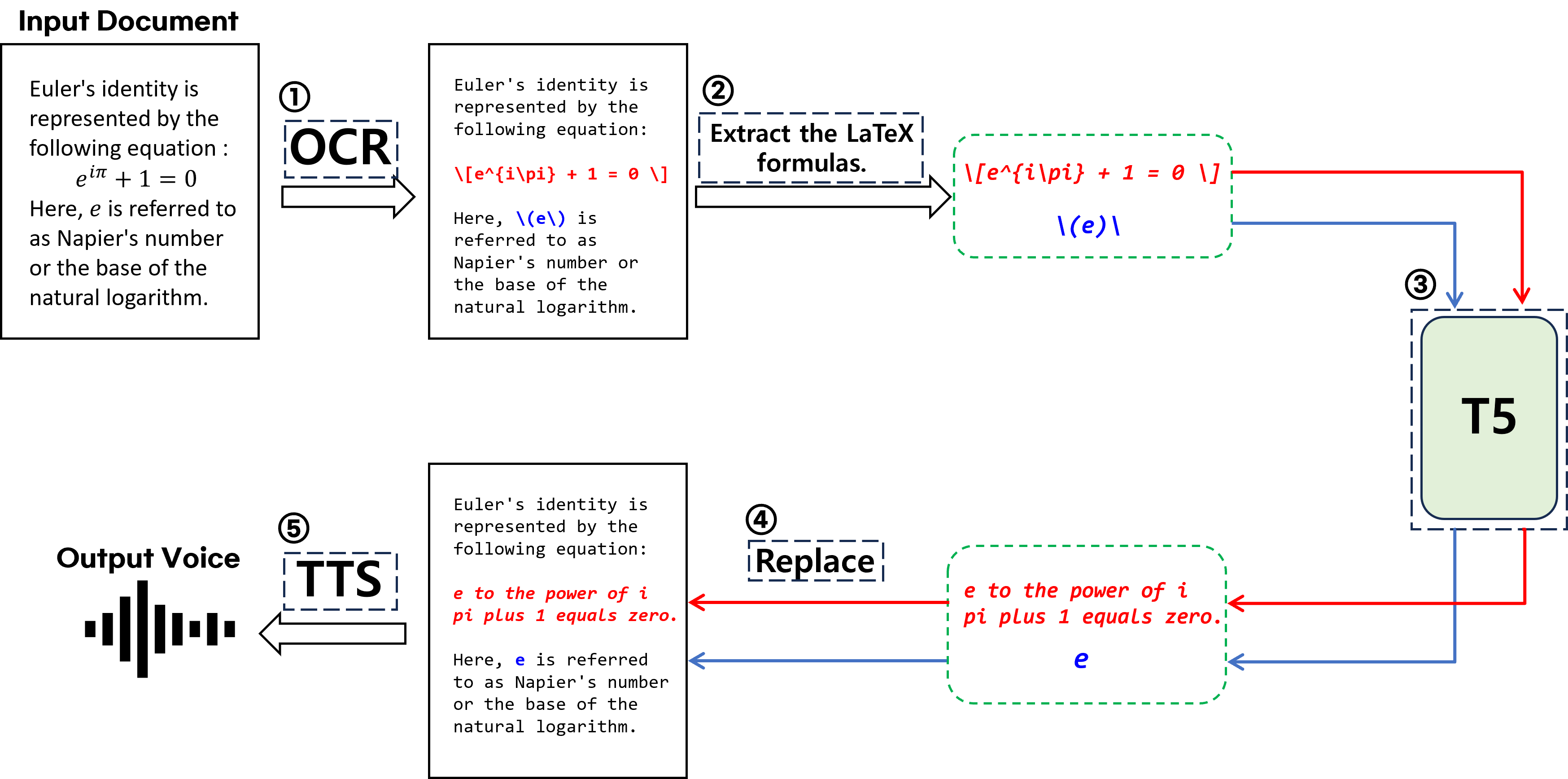} 
\caption{Our pipeline for reading the document correctly.}
\label{fig1}
\end{figure*}

To implement this process, we fine-tuned T5-small\cite{raffel2023exploringlimitstransferlearning}, one of the small language models, to create a \textit{LaTeX translator}, which identifies and converts the formulas from the OCR results into spoken English. We then implemented a pipeline that inputs the final text into the TTS system, with all formulas converted to spoken English. Our proposed MathReader can perfectly read not only general English, but also formulas, which will greatly assist visually impaired individuals in their learning. The contributions of this study are as follows:

\begin{itemize}
\item We identified that the \textit{read aloud} feature provided by existing software applications do not function well with formulas.
\item Our MathReader addresses this issue by integrating an OCR model, small-language model T5, and TTS model. Our pipeline can read all the formulas in documents without omissions.
\end{itemize}

\section{Related Works}

Most modern academic documents are written in LaTeX; in fields such as mathematics and science, where many formulas must be written, the use of LaTeX is essential. As a result, current Optical Character Recognition (OCR) technologies are being developed to convert academic documents into the LaTeX format, and advancements in deep learning-based neural network models\cite{wang2020pdf2latex, blecher2023nougatneuralopticalunderstanding, peng2023upocr, orji2023advancing, mishra2018sequence} have achieved remarkable success in this field. With OCR technology based on computer vision models, even older academic documents, which have only been preserved as images and were not originally created on computers, can be converted into LaTeX format.

However, editing or reading documents in LaTeX is extremely challenging for visually impaired individuals, which has led to the development of various assistive technologies\cite{sanmiguel2015latexvoice, 10.1145/3290605.3300734, 10.1145/3373625.3417009, jung2024texbleuautomaticmetricevaluate, kortemeyer2023using, yang2024latex, baker2021editing, hyeon2024mathspeechleveragingsmalllms}. One of the most prominent technologies is the conversion of LaTeX into speech using TTS (Text-to-Speech) technology. A study by \cite{sanmiguel2015latexvoice} manually mapped a specific LaTeX code to spoken English for conversion into speech. However, this approach faces challenges in handling numerous exception cases. Our research utilized a small language model to correctly read all the formulas written in LaTeX without exceptions. \cite{10.1145/3290605.3300734} and \cite{10.1145/3373625.3417009} used Microsoft’s TTS engine\cite{MicrosoftTTS} in a LaTeX editing program developed for the visually impaired. However, their program was designed to assist visually impaired users in creating and editing documents in LaTeX, and they did not develop a program for reading existing documents, especially older documents that only exist as images. \cite{kortemeyer2023using} conducted research using GPT-4\cite{achiam2023gpt} to make LaTeX content accessible to visually impaired individuals, but GPT-4 is too expensive, and its inference speed is too slow to be practical. We overcame this limitation by using a much smaller language model, T5-small.

\section{Method}

Our pipeline converts documents into voices through a 5-stage process. (See Figure 1).

\subsection{OCR}  
Optical Character Recognition (OCR) is a task that recognizes characters in a document and converts them into textual data. With the advancement of neural networks, various attempts have been made to apply deep-learning models to OCR tasks. We used Nougat-small\cite{blecher2023nougatneuralopticalunderstanding}, which is based on the hierarchical vision transformer\cite{dosovitskiy2021an}, Swin Transformer\cite{Liu_2021_ICCV}. This allowed us to convert PDF documents into mmd files composed of a markup language. 

Because Nougat accepts the PDF format as input, the document must be prepared in the PDF format.  Because our goal is to obtain voice output from the document, we can input the text from the mmd file generated by Nougat into the TTS model to obtain the corresponding voice. However, current TTS models cannot read special characters. In other words, the TTS model cannot read LaTeX codes, which are written using numerous special characters in the mmd file. Therefore, we modified these LaTeX codes to make them readable using the TTS model.

\subsection{Extract Formulas}

When Nougat converts the PDF into an mmd file, the mmd file contains both general English text and LaTeX codes representing the formulas. Nougat marks the LaTeX formulas in the mmd file using special characters like \textbackslash[ ~ ]\textbackslash  or 
\textbackslash( ~ )\textbackslash. Using these special characters, we can locate all LaTeX formulas within the mmd file.

\subsection{Fine-tuned T5}
T5 is a pre-trained language model for various text-to-text tasks\cite{raffel2023exploringlimitstransferlearning}. T5 has been pre-trained in different sizes and is widely used for translation tasks. According to previous research\cite{jung2024mathbridgelargecorpusdataset}, converting LaTeX into spoken English text can be considered a type of "translation" task. Therefore, we used a fine-tuned T5 model to convert all LaTeX formulas in the mmd file into spoken English. 

To fine-tune T5 for the LaTeX translation task, we needed a dataset composed of (LaTeX - Spoken English) pairs, and because such a dataset already exists publicly\cite{jung2024mathbridgelargecorpusdataset}, we used it. Additionally, considering that using a large language model would result in slow inference speeds and reduced practicality, we used the smallest version, T5-small. Our experiments confirmed that T5-small produces spoken English of sufficiently good quality.

\subsection{Replace LaTeX with Spoken English}
Once we converted all the LaTeX formulas into spoken English using T5, we replaced the LaTeX formulas in the mmd file with spoken English equivalents, making the text readable by the TTS model.

\subsection{TTS}
After converting all LaTeX formulas into spoken English in Step D, there are no formulas in the mmd file that the TTS model cannot read. Therefore, the TTS model outputs a voice without any errors. We connected VITS\cite{pmlr-v139-kim21f}, a TTS model, to the pipeline. 

The algorithm for implementing our proposed MathReader is as follows. (See Algorithm 1)

\begin{algorithm}[h]
\caption{MathReader Algorithm}
\begin{algorithmic}[1]
\Require A document (PDF format) 
\Ensure Voice (wav format) 

\State Let $pdf$ be the input document. 
\State $mmd \gets$ OCR($pdf$) 
\State $latex\_patterns \gets$ $\backslash[$ $\sim$ ]$\backslash$ or $\backslash($ $\sim$  
 $)\backslash$
\Function{SeperateLaTeX}{$mmd$}
    \State $sentences \gets$ split $mmd$ using $latex\_pattern$
    \State \Return $sentences$
\EndFunction
\Function{translateLaTeX}{$sentences$}
    \State $replaced\_list \gets$ empty list
    \For{each $sentence$ in $sentences$}
        \If{$sentence$ matches $latex\_pattern$}
            \State $Spoken\_Eng \gets T5(sentence)$
            \State Append $Spoken\_Eng$ to $replaced\_list$
        \Else
            \State Append $sentence$ to $replaced\_list$
        \EndIf
    \EndFor
    \State \Return $replaced\_list$
\EndFunction 
\State $Seperated\_mmd \gets$ SeperateLaTeX($mmd$) 
\State $final\_text\_list \gets$ translateLaTeX($Seperated\_mmd$) 
\State $Voice \gets$ TTS($final\_text\_list$) 
\State \textbf{return} $Voice$
\end{algorithmic}
\end{algorithm}

\section{Experiments}

\subsection{Test Dataset}

We collected documents available as open source for testing. However, while we could access the LaTeX sources of these documents, we could not find a dataset with the corresponding spoken English transcriptions. Therefore, we manually typed the LaTeX formulas in the documents in spoken English. Our dataset was made publicly available to anyone for access.

\subsection{Training Setup}

We used NVIDIA H100 for training T5. The model was trained for 20 epochs, and the model with the lowest validation loss was selected. The learning rate started at 1e-4 and was adjusted during training using a linear learning rate scheduler. The batch size was set to 48 and both the input and output sequence lengths were set to 325.

\subsection{Experiment method}

We first saved the wav files for the test dataset documents using our MathReader and the comparison TTS document readers, Microsoft Edge and Adobe Acrobat. Then, using Naver Clova Note, a Speech-To-Text (STT) platform\cite{clovanote}, we converted all the spoken results into text and measured the Word Error Rate (WER) against the Ground Truth. In addition, owing to the nature of mathematical expressions, errors occur during speech-to-text conversion because of homophones. For example, \textit{y} is sometimes outputted as \textit{why}, or \textit{T} as \textit{Tee}. Considering this, we also measured the Character Error Rate (CER).

\subsection{Experiment Results and Analysis}

\begin{figure*}[t]
\centering
\includegraphics[width=\textwidth]{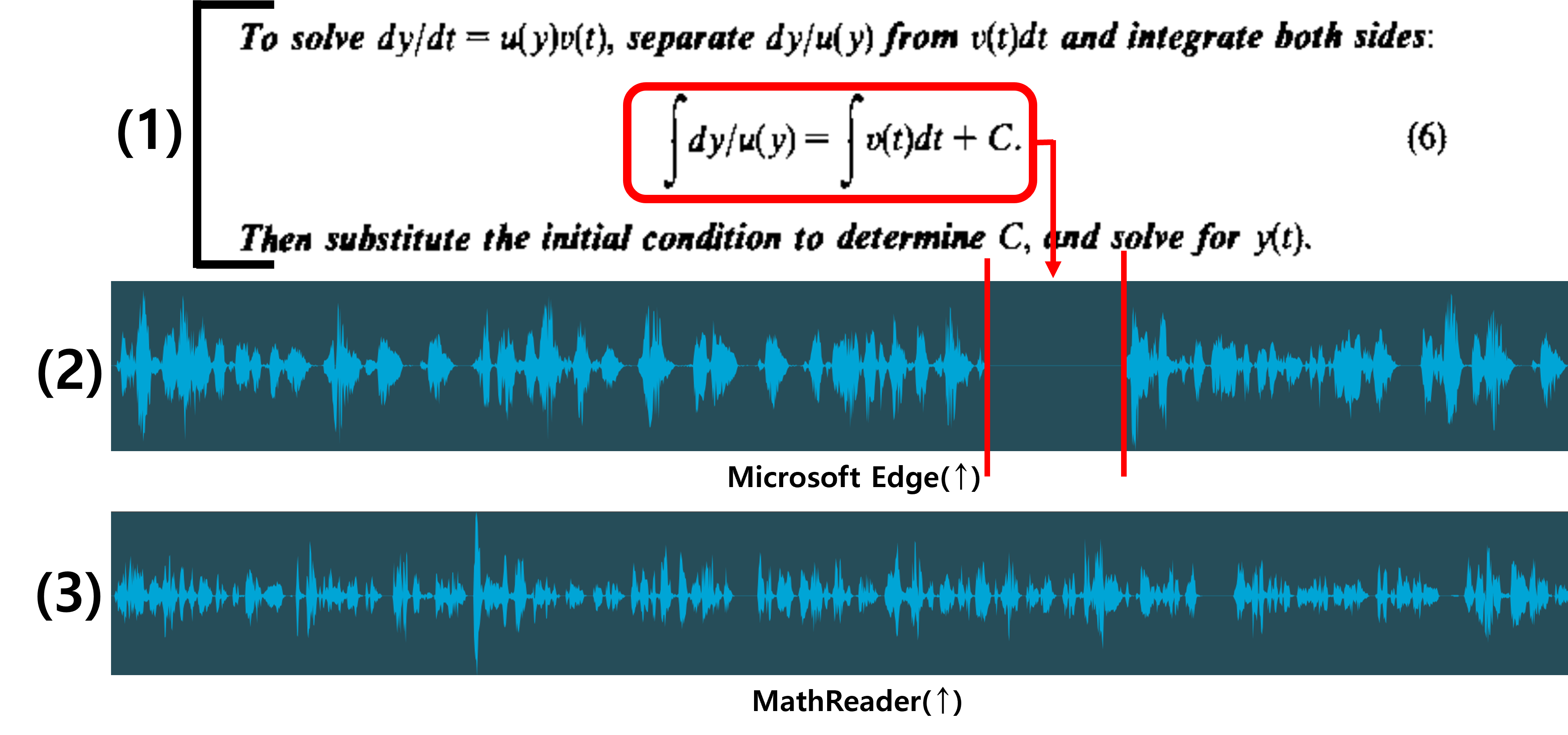} 
\caption{This figure shows an example of a TTS document reader skipping a formula from a document. \textbf{(1)} is a part of the document we used for testing. Because the document is old and of low quality, Microsoft Edge skips reading it aloud (See \textbf{(2)}). However, MathReader reads this correctly (See \textbf{(3)}). }
\label{fig2}
\end{figure*}

\begin{table}[h]
\centering
\caption{Experiment Results}
\begin{tabular}{@{}p{4cm}|p{2.02cm}|p{2.02cm}@{}}
    \toprule
    TTS document reader  & WER(\textdownarrow) & CER(\textdownarrow) \\
    \midrule
    \textbf{MathReader} & \textbf{0.281} & \textbf{0.148}  \\
    Microsoft Edge\cite{microsoftedge} & 0.510 & 0.341  \\
    Adobe Acrobat\cite{adobe}  & 0.617 & 0.454  \\
    MathReader without T5 & 0.663 & 0.491  \\
    \bottomrule
\end{tabular}
\label{table2}
\end{table}

The experimental results are listed in Table 2. MathReader achieved the lowest WER and CER compared with the other TTS document readers. Microsoft Edge showed the second-lowest error rates. The reason for the higher error rates in Microsoft Edge and Adobe Acrobat is that they read the text in the order it is recognized without considering the mathematical meaning of the formulas. An example of how each TTS document reader reads the formula is shown in Table 3.

\begin{table}[h]
\centering
\caption{An example of how each TTS document reader reads the formula}
\begin{tabular}{@{}p{1.88cm}|p{2.46cm}|p{3.7cm}@{}}
    \toprule
    Formula  & TTS document reader & Spoken English \\
    \hline
    \hline
     \multirow{3}{*}{$\sum\limits_{n=1}^{5}\left(\frac{1}{n}-\frac{1}{n+1}\right)$} & \textbf{MathReader}  & \textbf{Sum from n equals 1 to 5 of 1 over n minus 1 over n plus 1} \\ \cline{2-3}
     & Microsoft Edge & p 5, n equals 1, 1 n minus 1 n plus 1 \\  \cline{2-3}
    & Adobe Acrobat &  5, p, 11 n n+1, n=1 \\ 
    \hline
\end{tabular}
\label{table2}
\end{table}

Furthermore, when the document resolution was poor and formulas were not properly recognized, Microsoft Edge and Adobe Acrobat often skipped unrecognized formulas. An example is shown in Figure 2.  

For the ablation study, we conducted an experiment by removing T5-small from MathReader. We measured WER and CER using voices generated by directly inputting the OCR results into the TTS system. As a result, the LaTeX formulas in the document were either skipped or output inaccurately, leading to the worst experimental outcomes.

\subsection{Time Performance}

\begin{table}[h]
\centering
\caption{Time taken for MathReader execution. The time was measured for one PDF page, and the values in the table represent the average across all the test data.}
\begin{tabular}{@{}p{4.02cm}|p{4.02cm}@{}}
    \toprule
    Step  & Average Time (sec) \\
    \midrule
    OCR  (Nougat-small) & 12.54   \\
    Seperate LaTeX & 0.01   \\
    Translate and Replace (T5-small) & 4.86  \\
    TTS (vits) & 6.21 \\
    \midrule
    Total & 23.62 \\
    \bottomrule
\end{tabular}
\label{table3}
\end{table}

Table 4 lists the time taken to run MathReader. We measured the time spent at each stage and total time required to convert a single document into speech. When tested in an environment with an Intel(R) Xeon(R) Platinum 8480+ and NVIDIA H100, it took an average of 23.62 seconds to generate speech for one page document. The more text contained in the document, the longer is the process. In our test dataset, the shortest generated output speech was 134 seconds in length. This demonstrates MathReader's fast speech-generation speed relative to the length of the output speech, indicating that it can be used for real-time document-reading services, provided that a GPU is available.

\section{Conclusion}
In this study, we propose MathReader, a pipeline designed to convert documents into voices more accurately. Composed of Nougat-small, T5-small, and VITS, MathReader addresses the issue of inaccurate formula reading that occurs in traditional TTS document readers, such as Microsoft Edge and Adobe Acrobat. MathReader demonstrated significantly lower WER and CER than other TTS document readers, and we confirmed that it can be used for real-time document reading services in a GPU environment. Our MathReader will significantly contribute to reducing the inconvenience faced by individuals who rely on speech to understand documents, especially those who are visually impaired.

\section*{Acknowledgment}
This work was supported in part by Institute of Information \& communications Technology Planning \& Evaluation (IITP) grant (RS-2024-00399936, RS-2021-II211343, IITP-2023-RS-2023-00256081, IITP-2024-RS-2024-00397085, IITP-2024-RS-2024-00441407), and the New Faculty Startup Fund from Seoul National University. Samsung Memory Research Center (SMRC) provided research facilities for this work. J. Do is with ASRI, Seoul National University.

\bibliographystyle{unsrt}
\bibliography{refs}

\end{document}